# Synthesis of Hopfield Neural Network: Novel Results:

Garimella Rama Murthy,

Professor, Department of Computer Science,

Mahindra University, Bahadurpally, Hyderabad, India

**ABSTRACT**

Using the logical basis of synthesizing Hopfield Neural Network with desired corners of hypercube as stable states ( proposed in [1] ), it is proved that more corners of hypercube can be programmed as stable states ( whether the number of neurons is even or odd ). The research paper presents a new perspective to the so called "Programming Problem" of Hopfield Neural Network.

## 1. Introduction:

In an effort to realize some of the functions performed by biological neural networks, Artificial Neural Networks (ANNs) were proposed. The theory of ANNs led to the practically useful networks of neurons such as the Single Layer Perceptron (SLP), Multi-Layer Perceptron (MLP) etc. Hopfield, in an effort to model biological memories proposed the so called Hopfield Neural Network (HNN) which acts as an associative memory (HAM) [2].Thus Hopfield Associative Memory (HAM) was subjected to extensive research efforts and several interesting results were discovered.

One of the most important problems related to HAM was the so called "Synthesis Problem".This problem deals with synthesizing a HNN with certain states as the "desired memory states". Hopfield proposed a solution this problem which remained as a promising solution for a long time. The author understood the logical basis of Hopfield's synthesis procedure and provided novel solution to the problem. On careful examination of the method presented in [3], the author arrived at interesting new results reported in this research paper.

This research paper is organized as follows. In Section 2, related research literature is reviewed. In Section 3, new results related to synthesis of Hopfield Neural Network with stable states is proposed.

## 2. Review of Related Research Literature:

Hopfield neural network is a homogeneous nonlinear dynamical system associated with a network of McCulloch-Pitts artificial neurons captured by a graph based connectivity structure. Formally, consider a weighted graph with "N" neuronal nodes and the symmetric weight matrix ( called synaptic weight matrix ), $\bar{W}$ representing an Artificial Neural Network (ANN). Each neuron is in state '+1' or '-1' at any time 'n'. Thus, the state space of network of neurons is the unit symmetric hypercube, H i.e. H is the set of all N dimensional vectors with elements being '+1' or '-1'. Let the state vector ( of all N neurons ) at time 'n' be denoted by $\bar{V}(n)$. Thus, the dynamics of such homogeneous ( no external input ) nonlinear dynamical system based on the initial condition $\bar{V}(0)$ is specified by the following modes of operation.

- Serial Mode: State of only one neuron is updated at any time i.e.

$$v_i(n+1) = Sign\left\{\sum_{i=1}^{N} w_{ij} v_j(n) - t_i\right\} \; for\; n \geq 0.$$

- Fully Parallel Mode:

$$\bar{V}(n+1) = Sign\{\bar{W}\,\bar{V}(n) - \bar{T}\}\; for\; n \geq 0.$$
$$\bar{T} = (t_1, t_2, \ldots, t_N), \quad is\; the\; threshold\; vector.$$

- Other Parallel Modes: State updation takes place at more than one node, but strictly less than 'N' nodes.

In the state space of such a dynamical system, there are special states, called STABLE STATES. A stable state $\bar{U}$ is one such that

$$\bar{U} = Sign\{\bar{W}\,\bar{U} - \bar{T}\}.$$

The dynamics of such a nonlinear dynamical system is captured by the following convergence Theorem.

- Convergence Theorem:

  (i) The Hopfield Neural Network, starting in any initial condition always reached a stable state in the serial mode of operation if all the diagonal elements of $\bar{W}$ are all nonnegative.

  (ii) Also, in the fully parallel mode of operation, Hopfield Neural Network reaches a stable state or a cycle of length atmost 2.

Thus, the Hopfield Neural Network acts as an associative memory.

Hopfield's original work, formulated the problem of synthesizing an associative memory with desired stable states. He also provided a solution, using outer product of desired stable states ( that are corners of Unit Symmetric Hypercube ) to arrive at the synaptic weight matrix of Hopfield Associative Memory i.e.

$\bar{W} = \sum_{i=1}^{s} ( \bar{f}_i \; \bar{f}_i^T - I )$, where $\bar{f}_i's$ are corners of hypercube which constitutes the desired stable states.

Using Hopfield's synthesis procedure, Bruck et.al reasoned that such Hopfield Neural Network will have exponentially many "Spurious Stable States". Also, in [4], the author derived interesting results related to the stable states of a Hopfield Neural Network. The author's approach to the Hopfield Neural Network synthesis problem was based on linear algebraic structure of stable states of HAM. We realized that this approach is very promising. The author presented a formulation of "Optimal Synthesis" of HAM in [4]. This research paper complements the results derived in [4].

## 3. Linear Spaces Spanned by Eigenvectors: Stable States: Domains of Attraction:

The notion of "programmability" of certain corners of hypercube as "desired" stable states started with the so called "outer product synthesis" procedure of Hopfield. Understanding the logical basis of the Hopfield's procedure, a generalization was proposed in [1].

- It was proved in [1] that the corners of hypercube that are eigenvectors of the synaptic weight matrix corresponding to positive eigenvalues are stable states and the eigenvectors corresponding to negative eigenvalues are "anti-stable states" [5].

Even in the research papers [1], [4], it was not realized that it is possible to program MORE corners of hypercube as "desired stable states". In this section, using linear algebraic arguments, we investigate such a possibility.

We realize that the synaptic weight matrix, W of HNN is symmetric. Hence, various linear algebraic results can be invoked to derive new results related to the performance of HAM.

- From linear algebra of symmetric matrix, the null space of W and the linear space spanned by the eigenvectors associated with non-zero eigenvalues ( always real numbers ) are "orthogonal".
- We now specifically focus on the null space of W and investigate the nature of stable states reached ( in the serial mode of operation of HAM ) with the initial condition being a vector in the null space of W.

**Lemma 1:** Let $\bar{u}$ be any corner of unit symmetric hypercube which lies in the null space of W and which is utilized as the initial condition of HAM running in the serial mode of operation. The stable state reached ( by the HNN ) is always a specific vector $\bar{f}$ for any such $\bar{u}$ .

**Proof:** By the condition on $\bar{u}$, we have that Sign ( $\overline{W}\ \bar{u}$ ) = Sign ($\bar{0}$ ) = $\mp\bar{e}$, where $\bar{e}$ is a vector of "all ones" ( by consistent definition of signum function operating on the zero vector ).

By the convergence Theorem of Hopfield Neural Network, for all the vector $\bar{u}$ lying in the null space of W, in the serial mode of operation, same corner of hypercube $\bar{f}$ which is a stable state is reached Q.E.D

**Note:** Every vector in the null space of W is in the Domain of Attraction (DOA) of the stable state $\bar{f}$. Further with $\bar{x}$ , a corner of hypercube as the initial condition of HAM, a corner of hypercube $\bar{y}$ lying in the null space of W is transited, the HAM converges to the stable state $\bar{f}$.

The following Lemma deals with the corners of hypercube which are NOT in the null space spanned by eigenvectors corresponding to a repeated eigenvalue. We assume that the threshold vector is Zero vector i.e. $\bar{T} \equiv \bar{0}$.

**Lemma 2**: Corners of unit symmetric hypercube which are in the linear space spanned by eigenvectors corresponding to a repeated eigenvalue, are stable states, if the repeated eigenvalue is positive and are anti-stable states if the repeated eigenvalue is negative.

**Proof**: Let $\bar{f}_1, \bar{f}_2, \ldots, \bar{f}_r$ be eigenvectors corresponding to a positive valued eigenvalue, $\mu$ of algebraic multiplicity 'r'. Let the vector $\bar{z}$ lie in the linear space spanned by $\bar{f}_1$, $\bar{f}_2, \ldots, \bar{f}_r$ and is a corner of unit symmetric hypercube i.e.

$$\bar{f}_1, \bar{f}_2, \ldots, \bar{f}_r \qquad \bar{z} = \sum_{i=1}^{r} \alpha_i \bar{f}_i.$$

Also, let $$\bar{W} = \sum_{i=1}^{N} \alpha_i \bar{f}_i \ \bar{f}_i^T .$$

Since, the linear space spanned by $\bar{f}_1, \bar{f}_2, \ldots, \bar{f}_r$ is orthogonal to the linear space spanned by the eigenvectors corresponding to the other eigenvalues, we have that

$$\bar{W}\bar{z} = \sum_{i=1}^{r} \mu \, \alpha_i \bar{f}_i = \mu \left( \sum_{i=1}^{r} \alpha_i \bar{f}_i \right).$$

Hence Sign $(\bar{W}\bar{z}) = Sign(\mu \bar{z}) = \bar{z}$ if $\mu > 0$ and

Sign $(\bar{W}\bar{z}) = -\bar{z}$ $if$ $\mu < 0$.

Hence the claim of the Lemma follows.

**Corollary:** It readily follows that the corner of hypercube $\bar{z}$ is not orthogonal to the eigenvectors $\bar{f}_1, \bar{f}_2, \ldots, \bar{f}_r$ even if they are corners of unit hypercube. But $\bar{z}$ is orthogonal to the corners of hypercube which are in the linear space of eigenvectors corresponding to other repeated eigenvalues. In this sense, vectors like $\bar{z}$ are constrained as PROGRAMMED STABLE STATES.

**Note:** The above lemma doesnot make any assumption on $\bar{f}_1, \bar{f}_2, \ldots, \bar{f}_r$ being a Hadamard basis ( as considered in the research paper [4] ) or even $\bar{f}_1, \bar{f}_2, \ldots, \bar{f}_r$ being corners of the unit symmetric hypercube. This is a significant fact since, we proved in [4] that if N is odd, atmost one eigenvector can be a corner of unit symmetric hypercube.

**Significance of Above Lemma**: Hopfield's synthesis procedure

i.e. $\bar{W} = \sum_{i=1}^{s} ( \bar{f}_i \ \bar{f}_i^T - I )$ is based on the corners of unit symmetric hypercube being eigenvectors of $\bar{W}$ corresponding to the repeated eigenvalue 'N-s' ( of algebraic multiplicity 's', a positive integer less than N ). In [4], the author inferred that there is freedom in choice of eigenvalues in the proposed GENERALIZED SYNTHESIS approach. It was not clear how that freedom can be capitalized. *More interestingly, the above lemma enables programming corners of hypercube which are not eigenvectors of* $\bar{W}$ *as the stable states.* Specifically, such corners of the unit symmetric hypercube lie in the linear space spanned by eigenvectors corresponding to repeated eigenvalues. In view of this innovative idea, repeated eigenvalues ( of algebraic multiplicity larger than one ) can be grouped into different sets and correspond to orthogonal eigenvectors. As specified earlier, this approach of programming desired stable states can be utilized even when N is odd.

- As mentioned earlier, we assume that the threshold vector is a zero vector ( realizing well that by constraining eigenvalues, the results can be generalized to the more general case where the threshold vector is a non-zero vector. This generalization is very similar to that in [4] and is avoided for notational convenience and brevity ).

**Note:** In view of Lemma 2, the stable states lying in the linear spaces spanned by eigenvectors corresponding to different repeated eigenvalues are orthogonal ( as desired by the original Hopfield's synthesis ).

Now, we focus on the results related to "Domains of Attraction (DoA)" associated with the stable states.

By definition of stable state, we have that

$$\text{Sign}\ (\ \overline{W}\bar{f}\ ) = \bar{f} \text{ implies Sign}\ (\ \overline{W}\ (-\bar{f}\ )\ ) = -\bar{f}\ .$$

Hence the Domains of attraction of stable states are symmetrically located.

- In view of Lemma 2, certain domains of attraction can be divided into distinct sets based on "programmed/desired" stable states using eigenvectors associated with repeated eigenvalues. Specifically, we realize that if an initial condition $\overline{u_0}$ based HAM ( associated with a HAM running in serial mode ) transits through a corner of hypercube $\overline{x_0}$ which is in the Domain of Attraction (DoA) of a programmed stable state, $\bar{f}$, the succeeding states ( corners of hypercube ) are all in that DoA.

**Note:** Let $\overline{W}$ be singular. In view of Lemma 1, Lemma 2, the corners of unit hypercube that lie in the linear space spanned by eigenvectors corresponding to zero eigenvalue ( i.e. corners of unit hypercube in the null space of W ) are all in the DoA of $\bar{f}$ ( as specified in Lemma 1 ).